\documentclass[conference]{IEEEtran}
\usepackage{times}

\usepackage[numbers]{natbib}
\usepackage{multicol}
\usepackage[bookmarks=true,colorlinks=true]{hyperref}

\usepackage{graphicx}

\usepackage{booktabs}
\usepackage{multirow}

\usepackage{amsmath}

\newcommand{\ie}{\textit{i.e.}, } %
\newcommand{\eg}{\textit{e.g.}, } %
\newcommand{\mysubsection}[1]{\vspace{2mm}\noindent{\textbf{#1}}}
\DeclareMathOperator*{\argmax}{arg\,max}

\begin{document}

\title{Spatial Action Maps for Mobile Manipulation\vspace{-4mm}}

\author{
\authorblockN{Jimmy Wu$^1$, Xingyuan Sun$^1$, Andy Zeng$^2$, Shuran Song$^3$,\\
Johnny Lee$^2$, Szymon Rusinkiewicz$^1$, Thomas Funkhouser$^{1,2}$}
\authorblockA{$^1$Princeton University\quad$^2$Google\quad$^3$Columbia University}
\vspace{-10mm}
}

\maketitle

\begin{abstract}
Typical end-to-end formulations for learning robotic navigation involve predicting a small set of steering command actions (e.g., step forward, turn left, turn right, etc.) from images of the current state (e.g., a bird's-eye view of a SLAM reconstruction). Instead, we show that it can be advantageous to learn with dense action representations defined in the same domain as the state. In this work, we present ``spatial action maps,'' in which the set of possible actions is represented by a pixel map (aligned with the input image of the current state), where each pixel represents a local navigational endpoint at the corresponding scene location. Using ConvNets to infer spatial action maps from state images, action predictions are thereby spatially anchored on local visual features in the scene, enabling significantly faster learning of complex behaviors for mobile manipulation tasks with reinforcement learning. In our experiments, we task a robot with pushing objects to a goal location, and find that policies learned with spatial action maps achieve much better performance than traditional alternatives.
\end{abstract}

\IEEEpeerreviewmaketitle

\section{Introduction}

The prolific success of deep learning in vision~\cite{krizhevsky2012imagenet,szegedy2013deep,he2016deep,he2017mask} has inspired much recent work on using deep convolutional networks (ConvNets) for visual navigation and mobile manipulation in robotics.  In such an approach, ConvNets are trained to model a policy that maps from an agent's observation of the state (\eg images) to the probability distribution over actions (\eg steering commands) maximizing expected task success.

Most work in this domain has focused on methods to train the ConvNets: from learning to imitate human control strategies~\cite{muller2006off,bojarski2016end,gao2017intention,pfeiffer2017perception}, to discovering more complex long-term path planning behaviors with reinforcement learning~\cite{mnih2015human,zhu2017target,chen2018deep,shah2018follownet}.
Other works have considered the impact of different state representations: from forward-facing camera images~\cite{muller2006off,ross2013learning,bojarski2016end,pfeiffer2017perception} to top-down bird's-eye views (BEV) of the scene generated with Inverse Perspective Mapping (IPM)~\cite{gao2017intention,chen2018deep,shah2018follownet,bruls2019right}. However, there has been very little work on the impact of different action representations. Almost all navigation systems based on ConvNets consider only a small set of (possibly parameterized) egocentric steering commands (\eg move forward/backward, strafe left/right, rotate left/right, etc.)~\cite{ross2013learning,mnih2015human,gao2017intention,gupta2017cognitive,pfeiffer2017perception,zhu2017target,shah2018follownet,chen2018deep}.

\begin{figure}
\vspace{2mm}
\includegraphics[width=\columnwidth]{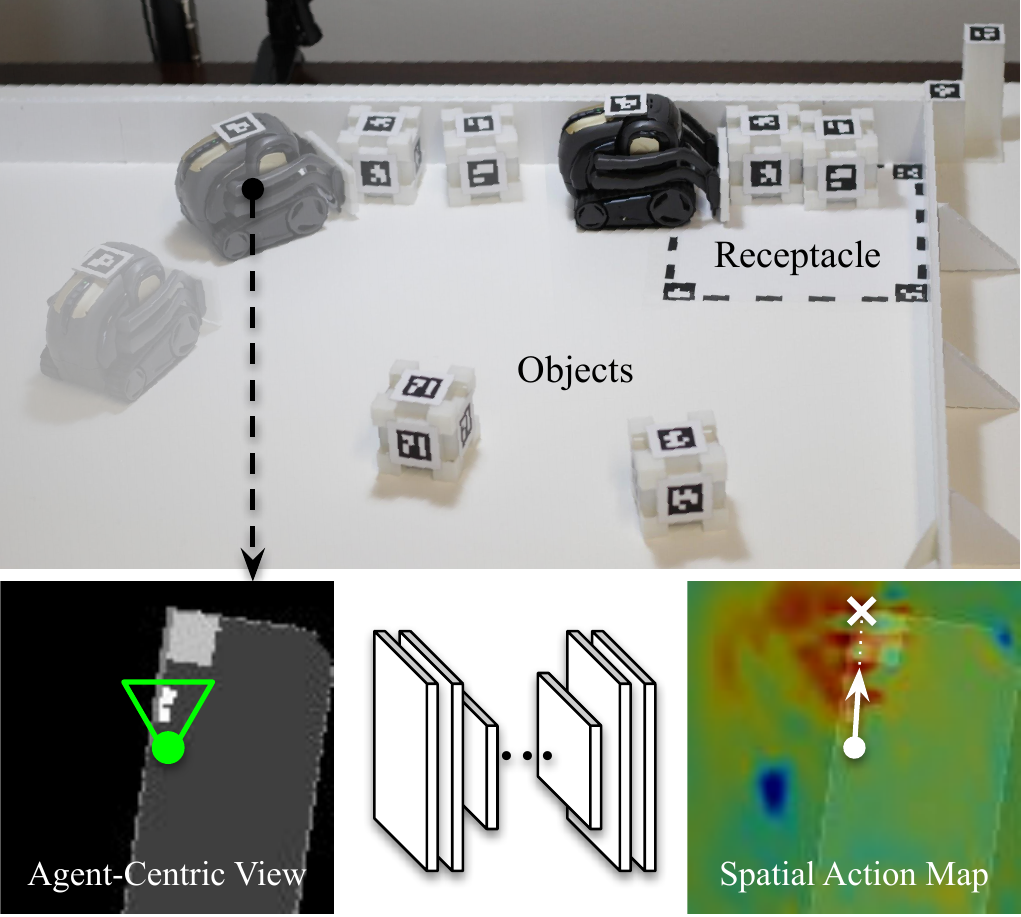}
\vspace{-6mm}
\caption{In this work, we propose ``spatial action maps'' as an action representation for mobile manipulation tasks.  In our setting, the goal is to train an agent to navigate around an unseen environment, find objects, and push them into a designated target receptacle.  It does so by iteratively predicting the Q-value of navigating to each location in a spatial action map (red is higher), and then selecting the best location to go to (marked with `x').}
\label{fig:teaser}
\vspace{-5mm}
\end{figure}

\begin{figure}
\includegraphics[width=\columnwidth]{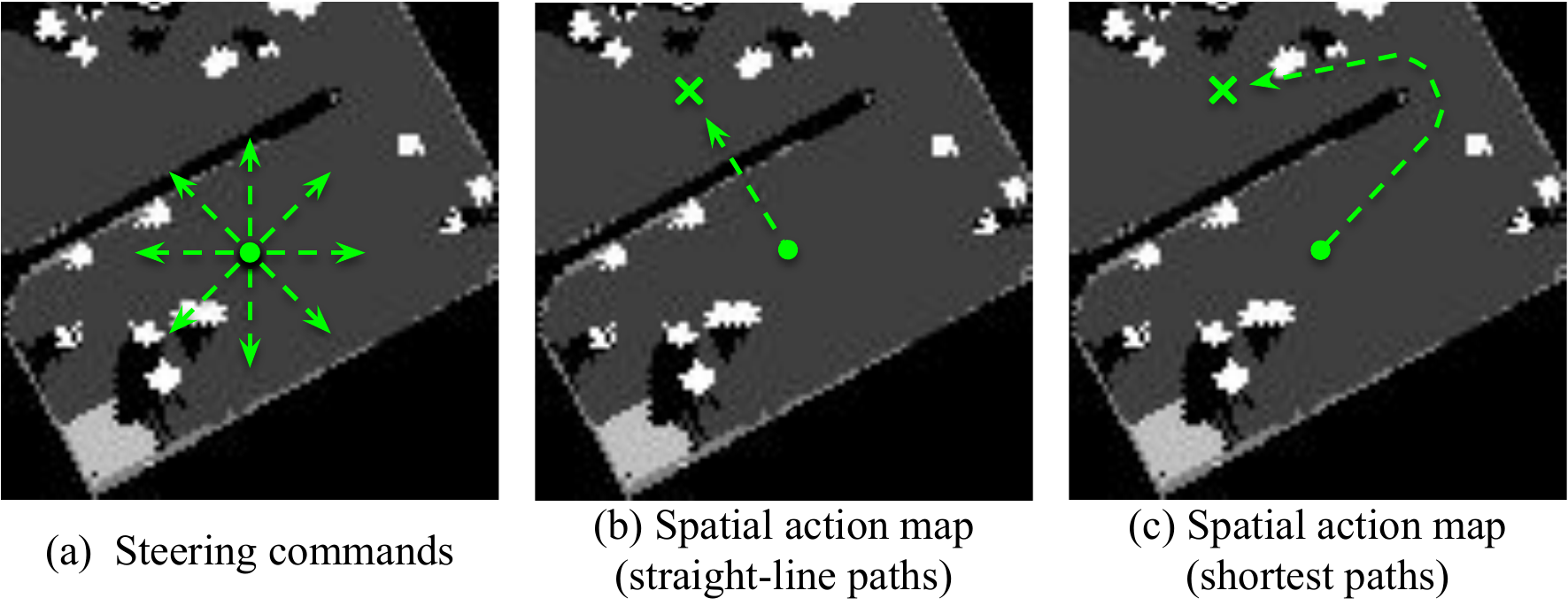}
\vspace{-7mm}
\caption{\textbf{Action representations.}  (a) shows all trajectories (green) available in a discrete set of steering commands.  The other images show example trajectories available in spatial action maps (every pixel is a potential trajectory endpoint).  We consider two variants: (b) where the agent follows a straight-line path to the selected target endpoint (denoted by  'x'), and (c) where it follows an estimated shortest path (our method).}
\label{fig:action-representations}
\vspace{-5mm}
\end{figure}

Representing actions with steering commands (Fig.~\ref{fig:action-representations}a) presents several problems for learning complex mobile manipulation tasks.
First, they are myopic, and thus each action can reach only a small subset of possible endpoints (\ie ones at a fixed distance and reachable by unobstructed straight-line paths).
Second, they each invoke only a small change to the state, and thus a long sequence of actions is required to make significant change (\eg navigate through a series of obstacles).
Third, they require a deep Q-network to learn a complex mapping from a high-dimensional input state representation (usually an image) to a low-dimensional set of (possibly parameterized) action classes, which may require many training examples.

In this work, we advocate for a new action representation, ``spatial action maps'' (Fig.~\ref{fig:teaser}). The main idea is to represent actions as a dense map of navigation endpoints: each action represents a move to an endpoint, possibly along a non-linear trajectory, and possibly with a task to perform there (Fig.~\ref{fig:action-representations}b-c).  The advantages of this action representation are three-fold.
First, the agent is not myopic -- it can move to any endpoint in the spatial action map with a single action~\cite{amato2014planning}, enabling goal-driven behaviors.
Second, each action can represent an arbitrarily complex navigation trajectory -- ours follow shortest paths to endpoints while avoiding obstacles (Fig.~\ref{fig:action-representations}c).
Third, it simplifies the mapping from states to actions -- in our system, each state is represented by an image of the reconstructed scene from a bird's-eye view (in IPM coordinates), and the action space is represented by an image representing the expected Q-value of navigating to every endpoint (also in IPM coordinates).  Since the state and action space lie in the same domain and are pixel-aligned, we can train a fully convolutional network (FCN) to map between them -- an efficient way to predict values for all possible actions in one forward pass of a network.

We study this action representation in the context of reinforcement learning for mobile manipulation, where an agent is tasked with exploring an unseen environment containing obstacles and objects, with the goal of navigating to push all objects into a designated target zone -- \ie pushing the objects into a receptacle, like a bulldozer (Fig.~\ref{fig:teaser}).  At every step, the only information available to the agent is an overhead image representing a partial reconstruction of its local environment (everything it has observed with a forward-facing RGB-D camera, transformed into an IPM representation and accumulated over time to simulate online SLAM/reconstruction).   The agent feeds the state image into an FCN to produce an action image, which encodes the Q-value of moving to every endpoint location along the estimated shortest path trajectory.  It executes the action with highest Q-value and iterates.

The spatial action map representation has the key benefit that the spatial position of each state-action value prediction (with respect to the input IPM view) represents a local milestone (trajectory endpoint) for the agent's control strategy.  We conjecture that it is easier for convolutional networks to learn the state-action values of these navigational endpoints (as opposed to abstract low-level steering commands), since each prediction is aligned and translationally anchored to the visual features of the objects and obstacles directly observed in the input map. This is motivated by gains in performance observed in other domains~-- for example, image segmentation~\cite{he2017mask} has been shown to benefit from pixel-aligned input and output representations, while end-to-end robotic manipulation~\cite{morrison2018closing,zeng2018robotic,zeng2018learning,zeng2019learning} is significantly more sample efficient when using FCNs to predict state-action values for a dense sampling of grasps aligned with the visual input. Using spatial action maps, our experiments show that we are capable of training end-to-end mobile manipulation policies that generalize to new environments with fewer than 60k training samples (state-action pairs). This is orders of magnitude less data than prior work~\cite{mnih2015human,zhu2017target}.

Our main contribution in this work is the spatial action map representation for mobile manipulation.
We investigate its use with a variety of action primitives (take a short step, follow the straight-line path, follow the shortest path) and state input channels (partial scene reconstructions, shortest path distances from the agent, shortest path distances to the receptacle).
In simulation environments, we conduct ablation studies and find that our proposed state and action representations give better performance and learning efficiency compared to alternative representations.
We show that our policies also work in real environments.
Supplemental materials, including code, simulation environments, and videos of our robot in action, are available at \url{https://spatial-action-maps.cs.princeton.edu}.

\begin{figure*}
\includegraphics[width=\textwidth]{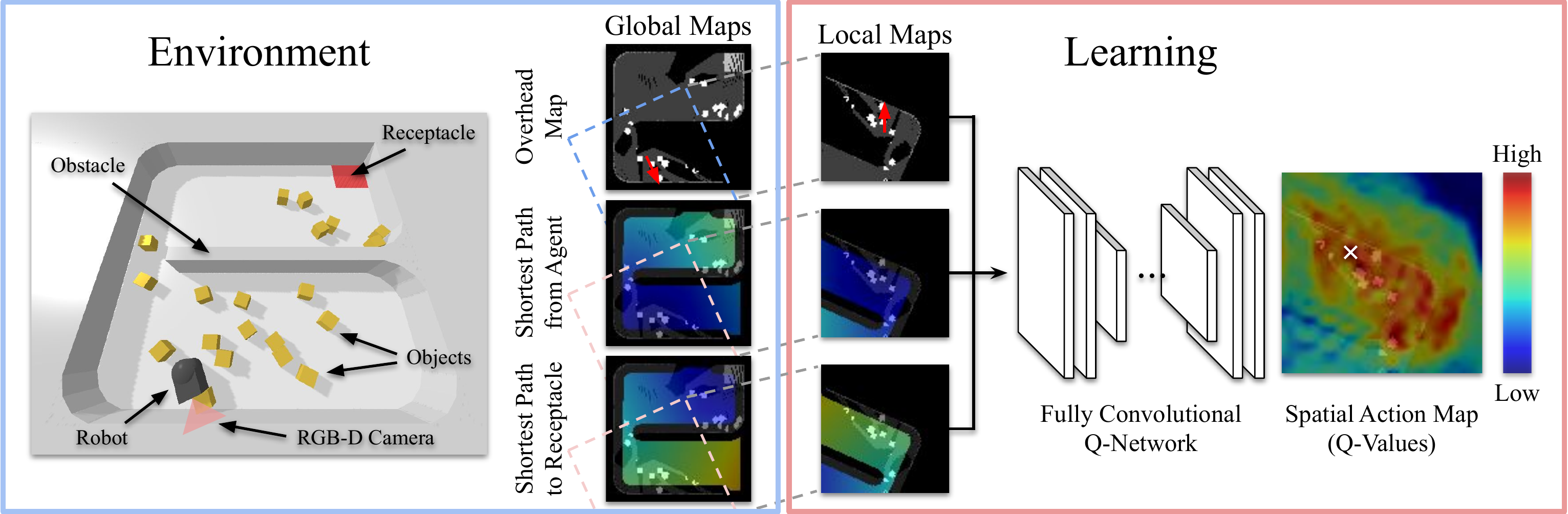}
\vspace{-6mm}
\caption{\textbf{Overview.} We train in a simulation environment, where we mimic online SLAM/reconstruction as the agent moves around. This is implemented via a virtual forward-facing RGB-D camera mounted on the robot. The robot incrementally builds up a global overhead map of the room, as well as an occupancy map used to compute single-source shortest path distances. Locally oriented and aligned crops of these maps are used to train our fully convolutional Q-network, which produces a dense, pixel-aligned spatial action map (Q-value map). Note that the robot position channel (see Fig.~\ref{fig:input-channels}) is omitted for clarity.}
\label{fig:overview}
\vspace{-5mm}
\end{figure*}
\section{Related Work}

\vspace{-2mm}\mysubsection{Navigation.}
There has been significant recent work on training agents to navigate virtual environments using first-person visuomotor control~\cite{kolve2017ai2,lowe2017multi,savva2017minos,anderson2018vision,wu2018building,xia2018gibson,yan2018chalet,savva2019habitat}. In a typical setup, the agent iteratively receives first-person observations as input (\eg images with color, depth, normals, semantic segmentations, etc.), from which it builds a persistent state (\eg a map), and selects one of many possible actions as output (\eg move, rotate, strafe, etc.) until it completes a task (\eg navigate to a given location, find a particular type of object, etc.). These works focus almost exclusively on how to best train neural networks for the task, for example using deep reinforcement learning~\cite{lowe2017multi,zhu2017target,chen2019learning}, supervised~\cite{tamar2016value,gupta2017cognitive} and self-supervised learning~\cite{qi2020learning}, or predicting the future~\cite{dosovitskiy2017learning}. They do not study how different parameterizations of network outputs (actions) affect the learning performance -- \ie the inputs are always in one domain (\eg images, GPS coordinates, etc.) and the outputs are in another (\eg move forward, rotate right, interact, etc.). In contrast, we investigate the advantages of dense predictions using spatial action maps, where the inputs and outputs are represented in the same spatial domain.

\mysubsection{Mobile Manipulation.}
While many navigation works assume a static environment, other works also consider environments with movable objects~\cite{stilman2007manipulation,van2009path,levihn2013hierarchical}.
In these works, the agent navigates to a goal location by pushing aside movable obstacles that are in the way.
For example, \citet{stilman2005navigation} propose the task of \textit{Navigation Among Movable Obstacles}, where the agent navigates in an environment that contains both static structures (\eg walls and columns) and movable obstacles.
While these methods do not assume a static environment, the task is similar -- the agent only needs to navigate to the goal location.
In contrast, in our task, the agent must come up with a navigational plan that will push all objects in the environment to the goal location, which is a much more complex problem.

Another line of work studies navigation for object manipulation~\cite{fuchikawa2005development,nishida2006development,zapata2018autotrans,lehner2018mobile}. These tasks (\eg picking and rearrangement of objects) are similar to ours, but the robot-object interactions considered in these tasks (\eg grasping) are often more predictable and happen over short time horizons, which makes it possible to apply simple heuristic-based algorithms to separately handle navigation and interaction.
In contrast, our setup requires long time-horizon robot-object interactions, which are less predictable and more difficult to plan.

\mysubsection{Dense Action Representations.}
Our work is inspired by recent work on dense affordance prediction for bin picking.
For example, the multi-affordance picking strategy of~\citet{zeng2018robotic} selects grasps by predicting a score for every pixel of an overhead view of a bin.
Similar approaches are used to predict affordances for grasping or suction in~\cite{zeng2018learning,zeng2019tossingbot,song2020grasping,zakka2019form2fit}.
However, these systems are trained in the more constrained setting of bin picking or assembly, where supervision is available for grasp success, and where motion trajectories to achieve selected grasps can be assumed to be viable and of equal cost.
In our work, we apply dense prediction to a more challenging scenario, where different actions have different costs (and some may not even be viable), and where long-term planning is required to perform complex action sequences.

\section{Methods}

In this paper, we propose a new dense action representation in which each pixel of a bird's-eye view image corresponds to the atomic action of navigating along the shortest path to the associated location in the scene.

To investigate this action representation, we consider a navigate-and-push setting in which an agent must explore an unseen environment, locate objects, and push them into a designated target ``receptacle'' (Fig.~\ref{fig:overview}).
This task may be seen as an abstraction of real-world tasks such as bulldozing debris or sweeping trash, and is sufficiently complex that it would be difficult to implement an effective hand-coded policy or learn a policy with traditional action representations.

Our agent is an Anki Vector, a low-cost mobile tracked robot approximately 10\,cm in length, augmented with a 3D-printed bulldozer-like end-effector (Fig.~\ref{fig:teaser}).
The objects to be pushed are 4.4\,cm cubic blocks, and the receptacle is located in the top right corner of a 1\,m by 0.5\,m ``room'' with random partitions and obstacles.
For ease of prototyping, we include fiducial markers on the robot and the objects, and track them with an overhead camera.
Our setting though, is intended to represent what would be possible with onboard sensing and SLAM.
Therefore, the only information made available to the agent is a simulation of what would be observed via an onboard forward-facing RGB-D camera with a 90$^\circ$ field of view, integrated over time with online mapping -- \ie our agents do not have access to ground truth global state.
This means that the agent must learn to act with partial or outdated information and actively seek out unexplored parts of the environment.

Our agents are trained in a PyBullet simulation~\cite{coumans2016pybullet}, where state observations are generated by rendering camera views of the environment.
We then execute learned policies in the real world, where the fiducial markers are used to update the state of the simulator (\eg robot and object poses).
Simulation enables us to train our agents over a wider range of environments than would be possible with physical robots, while sim-to-real mirroring enables us to evaluate the robustness and generalization of our policies to real-world dynamics.

In the following subsections, we provide details about our training setup and describe how the appropriate action representation proves crucial to improving the sample efficiency and generalizability of our policies.

\subsection{Reinforcement Learning (DQN) Formulation}

We formulate the navigate-and-push problem as a Markov decision process: given state $s_t$ at time $t$, the agent chooses to execute an action $a_t$ according to a policy $\pi(s_t)$, then transitions to a new state $s_{t+1}$ and receives a reward $r_t$. The goal of reinforcement learning is to find an optimal policy $\pi^*$ that selects actions maximizing total expected rewards $Q(s_t,a_t)=\sum_{i=t}^{\infty}\gamma^{i-t} r_i$, \ie a $\gamma $-discounted sum over an infinite horizon of future returns from time $t$ to $\infty$.

In this work, we investigate off-policy Q-learning~\cite{mnih2015human} to train a greedy policy $\pi(s_t)$ that chooses actions by maximizing a parameterized Q-function
$\argmax_{a_t} Q_\theta(s_t,a_t)$ 
(\ie state-action value function), where $\theta$ denotes the weights of our neural network (whose architecture we describe in Sec.~\ref{sec:method-training-details}).
We train our agents using the double DQN learning objective~\cite{van2016deep}. Formally, at each training iteration $i$, our objective is to minimize:
\vspace{-2mm}
\[\mathcal{L}_i = |r_t + \gamma Q_{\theta_i^-}(s_{t+1},\argmax_{a_{t+1}}{Q_{\theta_i}(s_{t+1},a_{t+1})})-Q_{\theta_i}(s_t,a_t)|\]
where $(s_t,a_t,r_t,s_{t+1})$ is a transition uniformly sampled from the replay buffer. Target network parameters $\theta^-$ are held fixed between individual updates. More training details are presented in Sec.~\ref{sec:method-training-details}.

\subsection{State Representation}
\label{sec:method-state-representation}

\begin{figure}
\includegraphics[width=\linewidth]{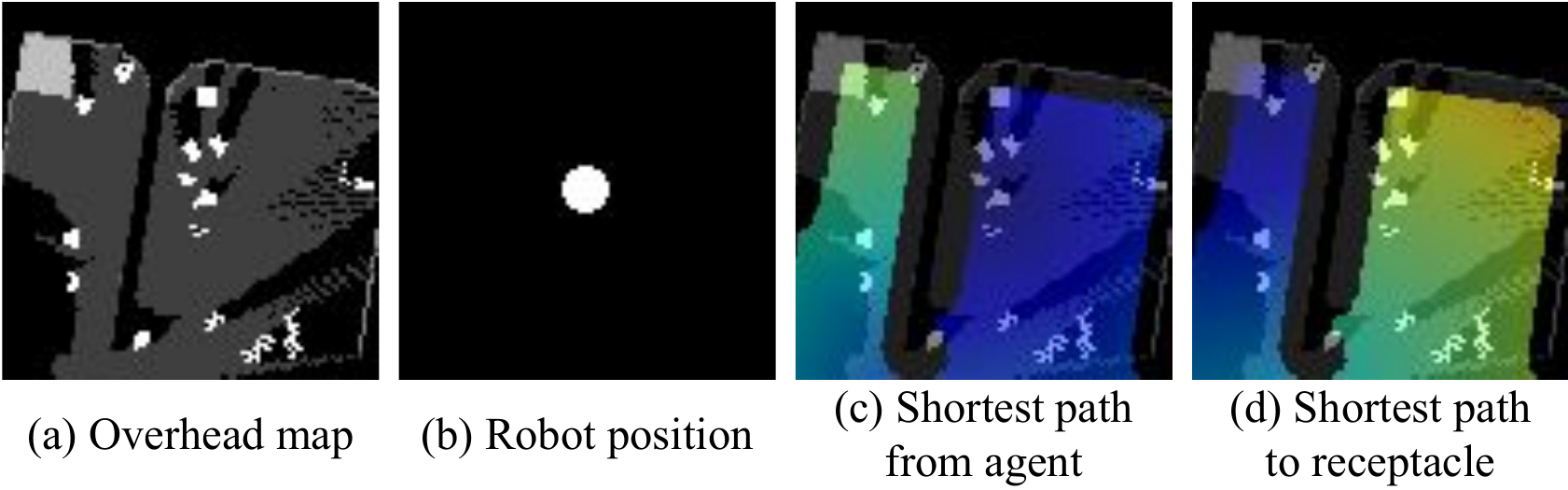}
\vspace{-7mm}
\caption{\textbf{Input channels.} From left to right, the pixel-aligned channels are all in the robot's local coordinate frame and correspond to (a) the observation, which is an overhead image of the reconstructed environment, (b) encoded robot position in local coordinates, which is always in the center of the image, (c) shortest path distance from the agent to each location in the image, and (d) shortest path distance to the receptacle from each location in the image. The receptacle is in the corner of the room, near the top left of these images. Color added for visualization only, green/yellow corresponds to higher values.}
\label{fig:input-channels}
\vspace{-5mm}
\end{figure}

Within our formulation, we represent the agent's observation of the state $s_t$ as a visual 4-channel image from a local bird's-eye view that is spatially aligned and oriented with respect to the robot's local coordinate frame (such that the robot is positioned in the center of each image and looking along the $y$ axis). This is similar to inverse perspective mapping (IPM), commonly used in autonomous driving~\cite{gao2017intention,chen2018deep,shah2018follownet,bruls2019right}. Each channel encodes useful information related to the environment~\cite{bhatti2016playing} (visualized in Fig.~\ref{fig:input-channels}) including: (1) a local observation of the robot's surroundings in the form of an overhead map, (2) a binary mask with a circle whose diameter and position encode the robot's respective size and location in the robot's coordinate frame, (3) an image where each pixel holds the shortest path distance from the agent to the corresponding location, and (4) an image where each pixel holds the shortest path distance to the receptacle from that location. The shortest path distances in the third and fourth channels are computed using an occupancy map generated only from local observations of obstacles (which are accumulated over time -- see next paragraph) and normalized so that they contain relative values rather than absolute. All unobserved regions are treated as free space when computing shortest path distances. This reflects a realistic setting in which a robot has access to nothing but its own visual observations, GPS coordinates, local mapping, and task-related goal coordinates.

\mysubsection{Online mapping.} Online SLAM/reconstruction is a common component of any real mobile robot system. In our experiments, this is implemented in the simulation using images from a forward-facing RGB-D camera mounted on the robot. The camera captures a new image at the end of every transition. Using camera intrinsics and robot poses, depth data is projected into a 3D point cloud, then fused with previous views to generate and update a global map of the environment. At the beginning of each episode in a new environment, this global map is initially blank. As the robot moves around in the environment, the global map fills in as it accumulates more partial views over time. This restriction encourages the agent to learn a policy that can explore unseen areas in order to effectively complete the task. Specifically, the robot gradually reconstructs two global maps as it moves around: (1) a global overhead map, which encodes objects and obstacles, and (2) a global occupancy map, used for shortest path computations in our high-level motion primitives as well as our state representations and partial rewards. The agent makes no prior assumptions about the layout of obstacles in the environment.

\begin{figure}
\includegraphics[width=\columnwidth]{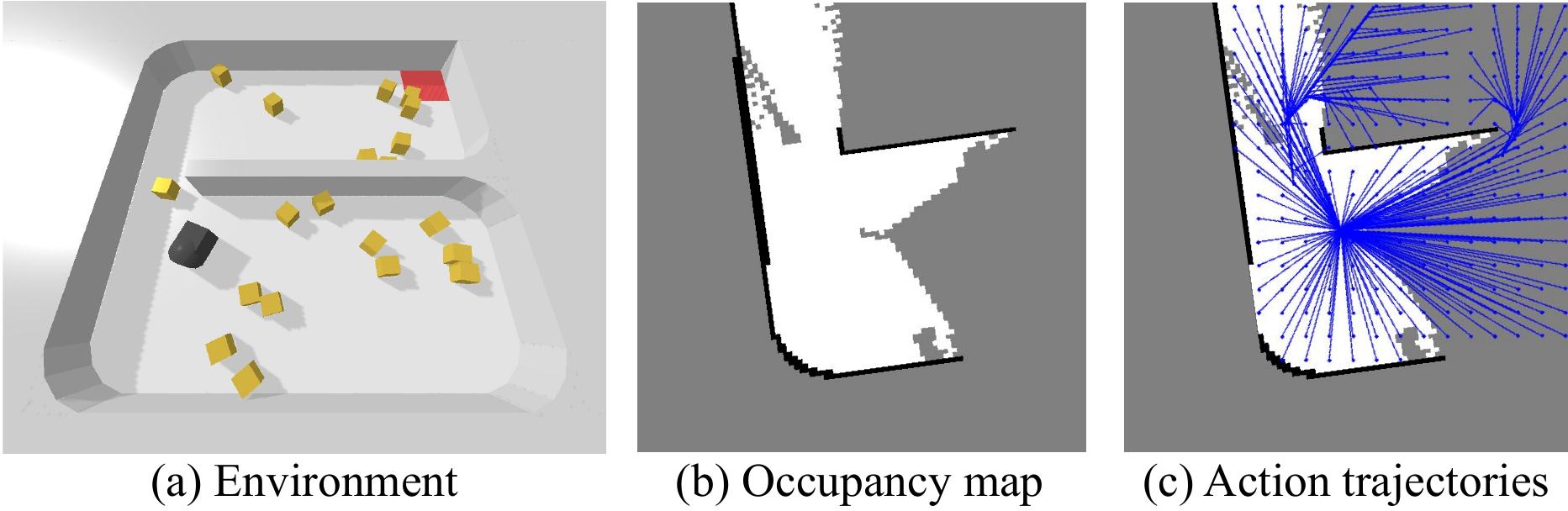}
\vspace{-7mm}
\caption{\textbf{Action space.} In our action space, which is spatially aligned with the local state representation, every pixel represents a move to the corresponding navigational endpoint along the shortest path trajectory. The robot is located in the left of the environment (a) and in the center of local maps (b) and (c). Shortest path trajectories (c) are estimated using an occupancy map (b) built up with online mapping. Note that the action space is pixelwise dense, but we only show a subset of paths in (c) for clarity.}
\label{fig:action-space}
\vspace{-5mm}
\end{figure}

\begin{figure*}
\begin{center}
\setlength\tabcolsep{0.3em}
\begin{tabular}{@{}cccc}
\includegraphics[width=0.24\textwidth]{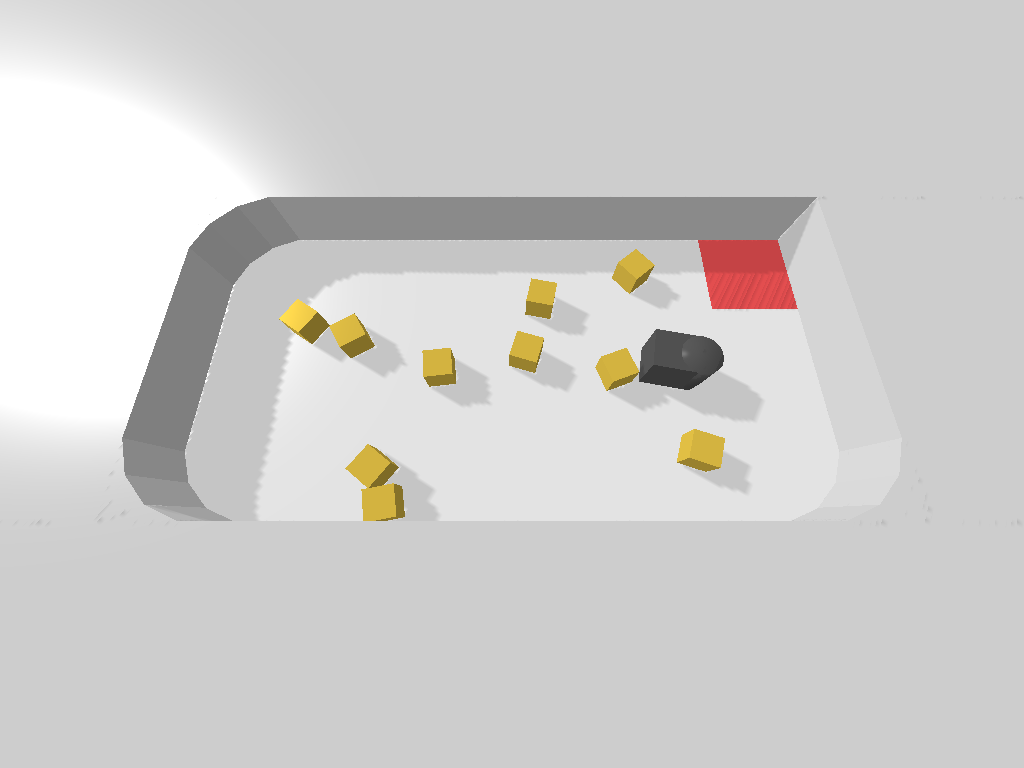} &
\includegraphics[width=0.24\textwidth]{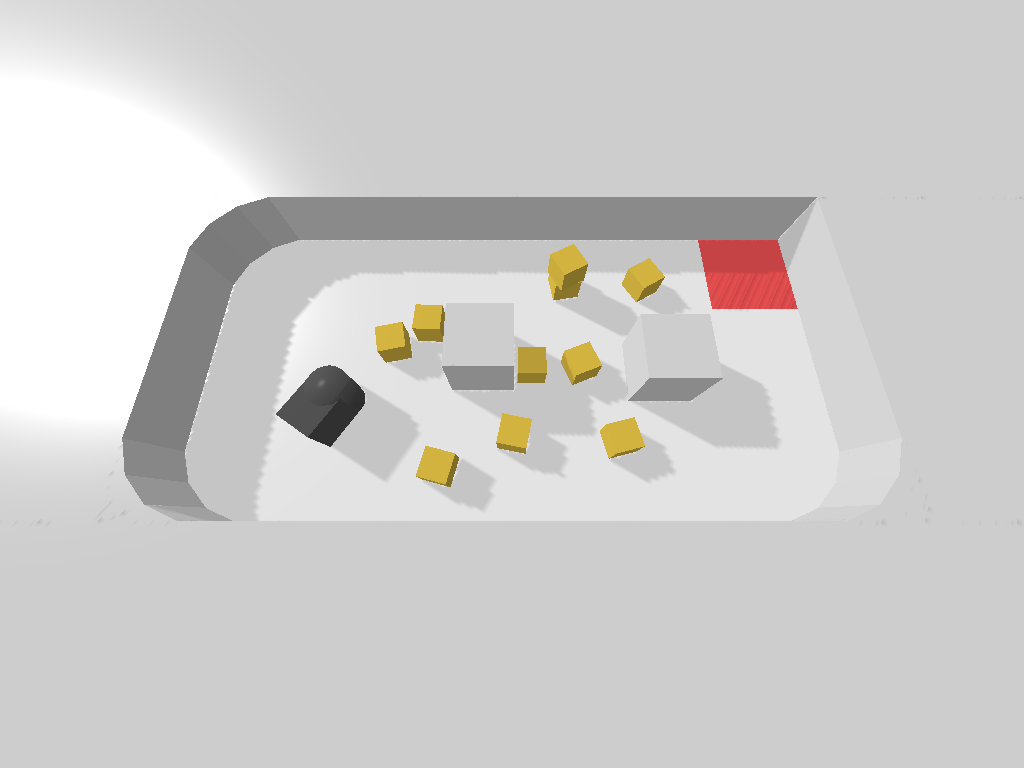} &
\includegraphics[width=0.24\textwidth]{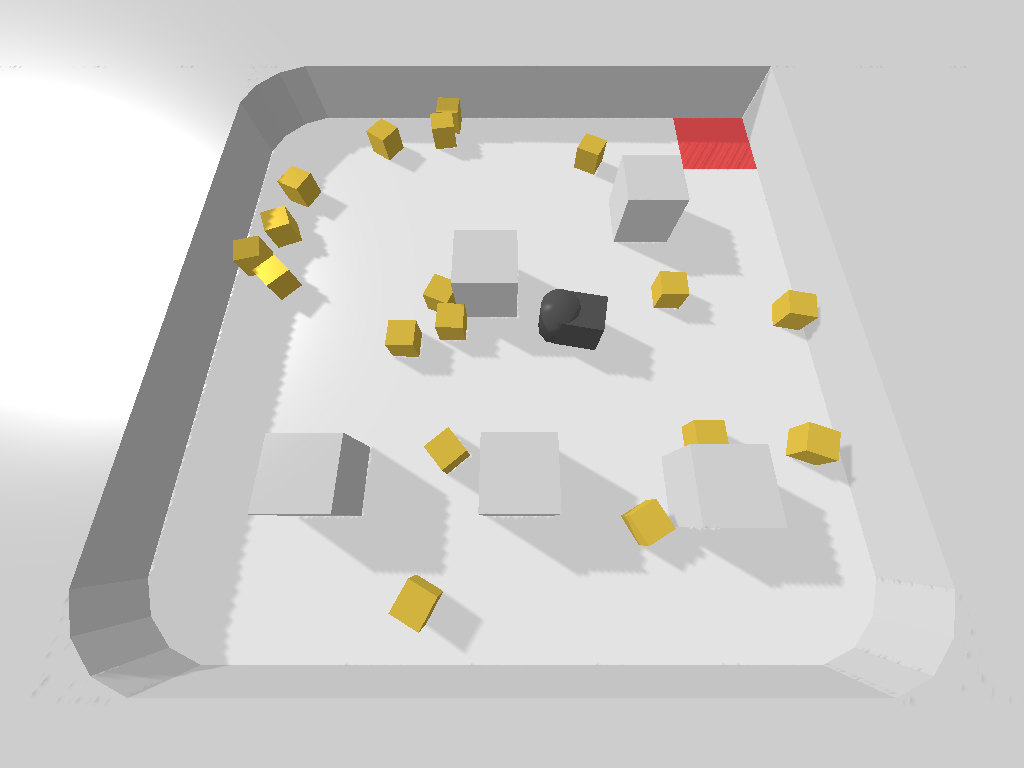} &
\includegraphics[width=0.24\textwidth]{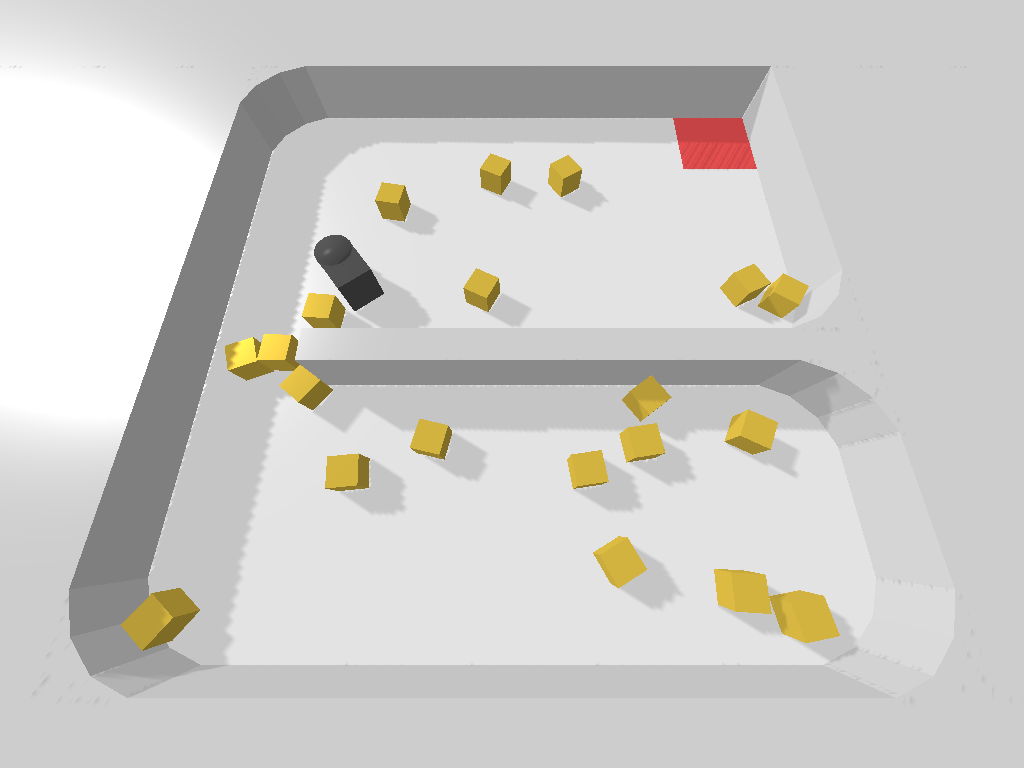} \\
\small{(a) SmallEmpty} & \small{(b) SmallColumns} & \small{(c) LargeColumns} & \small{(d) LargeDivider}
\end{tabular}
\end{center}
\vspace{-3mm}
\caption{\textbf{Environments.} We tested in four types of environments, each with a receptacle in the top right corner (red), randomized arrangements of objects (yellow blocks), and randomized placements of obstacles (gray) of increasing difficulty (left to right).}
\label{fig:environments}
\vspace{-5mm}
\end{figure*}

\subsection{Action Representation}
\label{sec:method-action-representation}

Our actions are represented by an image (\ie action map, illustrated in Fig. \ref{fig:action-representations}) identically sized and spatially aligned with the input state representation. Each pixel in this action map corresponds to a navigational endpoint in local robot Cartesian coordinates. At each time step, the agent selects a pixel location in the action map -- and therefore in the observed environment -- of where it intends to move to. Specifically, the selected location in the image indicates where the robot's forward-facing end effector should be located at after the action has been completed. The agent then uses a movement primitive to execute the move.

We experiment with two types of movement primitives: one that moves in a straight line towards the selected location, and one that follows the shortest path to the selected location. The straight line primitive simply turns the robot to face the selected location, and then moves forward until it has reached the location. The shortest path primitive (well-suited for environments with obstacles) uses the global occupancy map introduced in Sec.~\ref{sec:method-state-representation} to compute and follow the shortest path to the desired target location  (Fig.~\ref{fig:action-space}). For both, it is possible for the robot to collide with previously unseen obstacles, in which case a penalty (see Sec.~\ref{sec:method-training-details}) is incurred by the agent. Our experiments in Sec.~\ref{sec:experiments-simulation} compare this representation to discrete action alternatives (\eg steering commands) commonly used in the literature. Note that in empty environments without obstacles, these two movement primitives are equivalent.

\subsection{Network Architecture and Training Details}
\label{sec:method-training-details}

We model our parameterized Q-function $Q_\theta$ with a fully convolutional neural network (FCN), using a ResNet-18~\cite{he2016deep} backbone for all experiments. The FCN takes as input the 4-channel image state representation (described in Sec.~\ref{sec:method-state-representation}) and outputs a state-action value map (described in Sec.~\ref{sec:method-action-representation}). We removed the AvgPool and fully connected layers at the end of the ResNet-18, and replaced them with 3 convolutional layers interleaved with bilinear upsampling layers. The added convolutional layers use 1x1 kernels, and the upsampling layers use a scale factor of 2. This gives us an output that is the same size as the input. We also removed all BatchNorm layers from our networks, which provided more training stability with small batch sizes. To ensure that the FCN has an adequate receptive field, we designed our observation crop size (96 by 96) such that the receptive field of each network output covers over a quarter of the input image area, and thus always covers the center of the image in which the robot is located.

\mysubsection{Rewards.} Our reward structure for reinforcement learning (computed after each transition) consists of three components: (1) a positive reward of +1 for each object that ends up in the receptacle (objects in the receptacle are removed thereafter), (2) partial rewards and penalties based on whether each object was moved closer to or further away from the receptacle (proportional to the signed change in distance computed using either Euclidean distances or shortest path distances -- comparison in Sec.~\ref{sec:experiments-simulation}), and (3) small penalties of -0.25 for undesirable behaviors (collisions or nonmovement).

\mysubsection{Training details.} During training, our agent interacts with the environment and stores data from each transition
$e_t = (s_t, a_t, r_t, s_{t+1})$ in an experience replay buffer of size 10,000. At each time step, we uniformly at random sample a batch of experiences from the experience replay buffer, and train our neural network with smooth L1 loss (\ie Huber loss).
We pass gradients through only the single pixel in the input state that corresponds to the selected action for a transition \cite{zeng2018robotic,zeng2019learning}. We clip gradients such that they have magnitude at most 10. We train with batch size 32 and use stochastic gradient descent (SGD) with learning rate 0.01, momentum 0.9, and weight decay 0.0001. To account for varying distances traveled for different steps, we apply a discount factor $\lambda=0.99^{0.25 \cdot \text{dist}}$ with an exponent that is proportional to the distance traveled during that step.

In our experiments, we train for 60,000 transitions, which typically corresponds to several hundred episodes. Each episode runs until either all objects in the environment have been pushed into the target receptacle, or the robot has not pushed any objects into the receptacle for 100 steps. Our policies are trained from scratch, without any image-based pretraining~\cite{yen2020learning}. The target network is updated every 1,000 steps. Training takes about 9 hours on a single NVIDIA Titan Xp GPU.

\mysubsection{Exploration.} Before training the network, we run a random policy for 1,000 steps to fill the replay buffer with some initial experience. Our exploration strategy is $\epsilon$-greedy with initial $\epsilon=1$, annealed over training to 0.01 after 6,000 transitions.

\section{Experiments}

To test the proposed ideas, we run a series of experiments in both simulation and real-world environments.  We first describe the simulation experiments, which are used to investigate trade-offs of different algorithms, and then we test our best algorithm on the physical robot.

\mysubsection{Task.}
In every experiment, the robot is initialized with a random pose within a 3D environment enclosed by walls.  Within the environment, there is a set of cubic objects scattered randomly throughout free space and a 15\,cm by 15\,cm square receptacle in the top right corner, which serves as the target destination for the objects. The robot's task is to execute a sequence of actions that pushes all of the objects into the receptacle. Objects are removed from the environment after they enter the receptacle.

\mysubsection{Environments.}
We ran experiments with four virtual environments of increasing difficulty  (Fig.~\ref{fig:environments}). The first (SmallEmpty) is a small rectangular environment (1\,m by 0.5\,m) containing 10 randomly placed objects. The second (SmallColumns) adds a random number (1 to 3) of square (10\,cm by 10\,cm) fixed obstacles (like ``columns'') placed randomly. The third (LargeColumns) is a larger version (1\,m by 1\,m) with more columns (1 to 8) and more objects (20). The fourth (LargeDivider) replaces the columns with a single large divider that is fixed at a randomly chosen $y$ coordinate and spans 80\% of the $x$ dimension -- this last environment requires the robot to plan how to get from one side of the divider to the other by going through the narrow open gap, and thus is the most difficult.

\begin{figure}
\includegraphics[width=\linewidth]{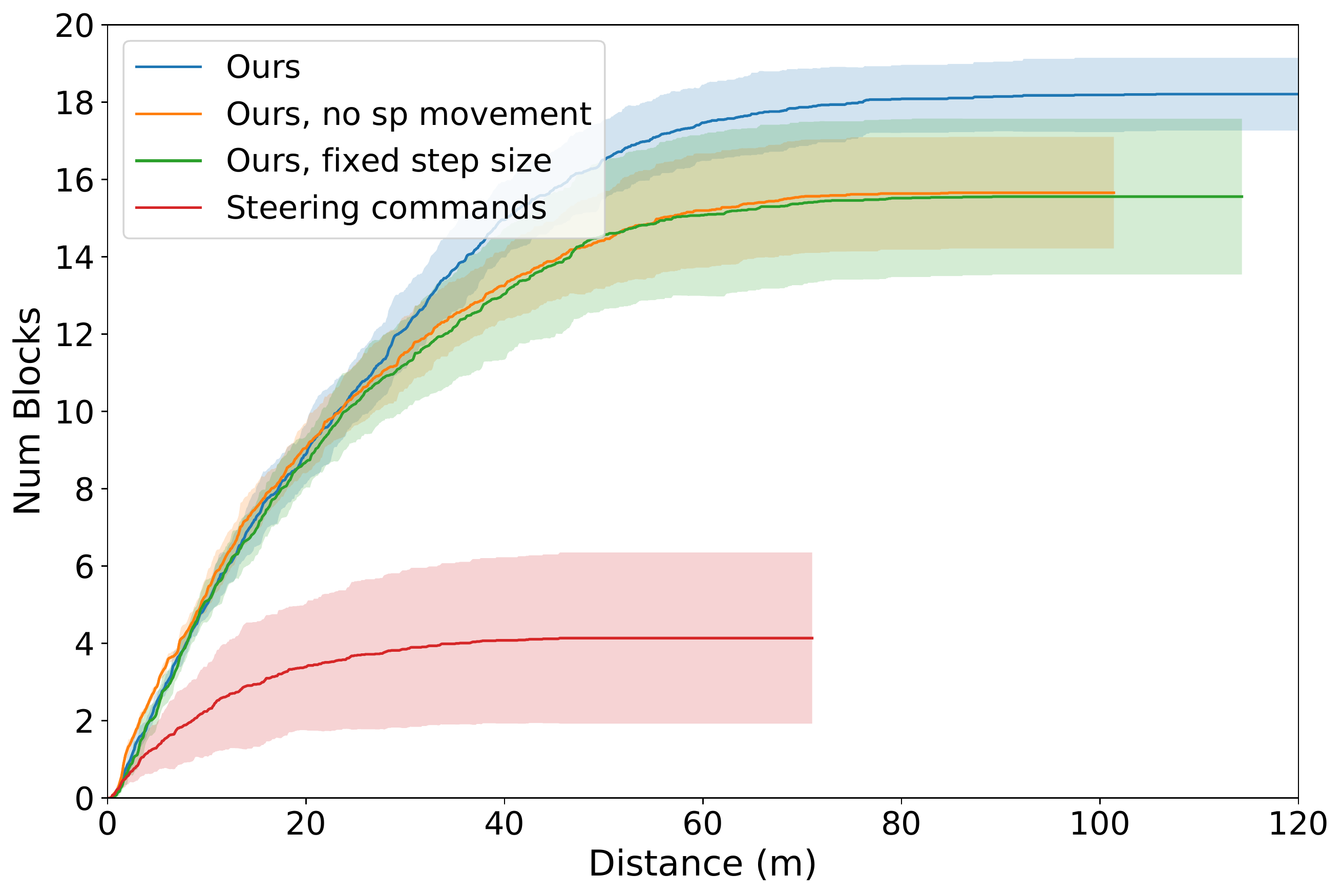}
\vspace{-8mm}
\caption{Number of objects pushed into the receptacle as a function of robot travel distance. These curves show evaluation of experiments that were trained with the LargeDivider environment (shown are means with shaded bars for standard deviations). Our method (blue) is slightly better than its ablation variants with more restricted actions (orange and green), and significantly better than the baseline method trained using steering commands (red).}
\label{fig:curves-eval}
\end{figure}

\begin{table}
\caption{Effect of Action Representation}
\label{tab:action-representations}
\begin{center}
\setlength\tabcolsep{0.4em}
\def\x{\hphantom{1}}
\begin{tabular}{l|cccc}
\toprule
\multirow{2}{*}{Environment} & \multirow{2}{*}{Ours} & \!\!No shortest path\!\! & Fixed & Steering \\
 & & movement & step size & commands  \\
\midrule
SmEmpty   & \x \bf{9.91} $\pm$ 0.11 &        n/a         & \x 9.75 $\pm$ 0.20 & 1.38 $\pm$ 0.20 \\
SmColumns & \x \bf{9.18} $\pm$ 0.14 & \x 7.88 $\pm$ 0.70 & \x 9.05 $\pm$ 0.38 & 0.82 $\pm$ 0.33 \\
LgColumns &   \bf{18.29} $\pm$ 0.45 &   14.70 $\pm$ 1.52 &   17.52 $\pm$ 0.82 & 1.20 $\pm$ 0.64 \\
LgDivider &   \bf{18.23} $\pm$ 0.92 &   15.66 $\pm$ 1.44 &   15.56 $\pm$ 2.01 & 4.14 $\pm$ 2.21 \\
\bottomrule
\end{tabular}
\end{center}
\begin{center}
(Number of objects pushed into the receptacle per episode)
\end{center}
\vspace{-7mm}
\end{table}

\mysubsection{Evaluation metrics.}
We evaluate each model by running the trained agent for 20 episodes in the environment it was trained on. Since the environments are randomly generated every episode, for each model, we set the random seed so that the exact same set of generated environments are presented to each model, including the initial robot pose, object poses, and obstacle placements. For all experiments, we do 5 training runs with the same setup and report the mean and standard deviation across the 5 runs. Trained agents are evaluated using an $\epsilon$-greedy policy with $\epsilon=0.01$.

\begin{figure}
\includegraphics[width=\columnwidth]{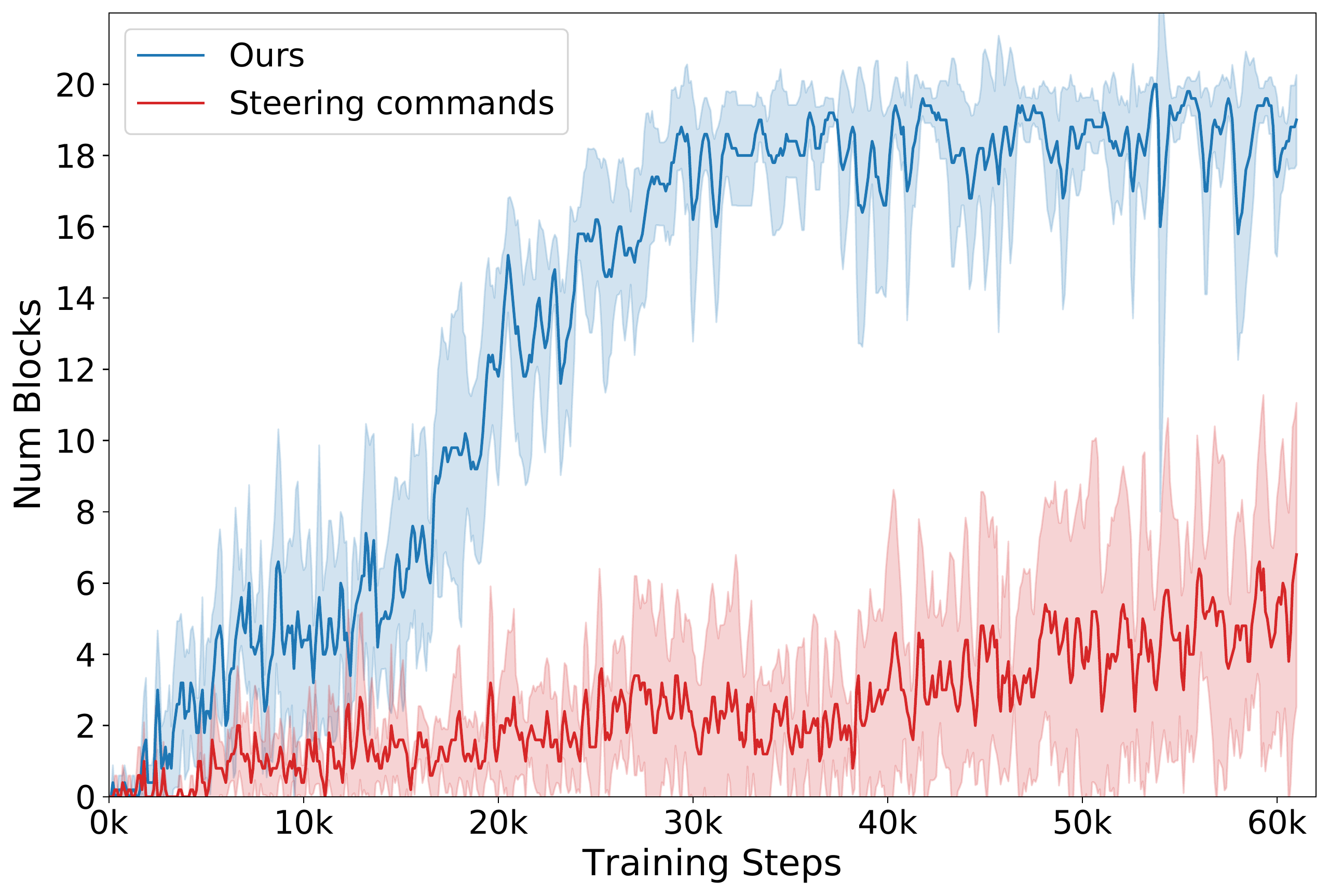}
\vspace{-8mm}
\caption{Training curves for the LargeDivider environment. Our agent trained with spatial action maps (blue) is significantly more sample efficient compared to the baseline that uses steering commands (red).}
\label{fig:curves-train}
\vspace{-5mm}
\end{figure}

We use two evaluation metrics. The first simply measures the number of objects that have been pushed into the receptacle at the end of an episode (Tab.~\ref{tab:action-representations}). The second plots the number of objects successfully pushed into the receptacle as a function of how far the robot has moved (Fig.~\ref{fig:curves-eval}). Additionally, we compare training sample efficiency by plotting objects per episode on the training environments, as a function of training steps (Fig.~\ref{fig:curves-train}). Higher numbers are better.

\subsection{Simulation Experiments}
\label{sec:experiments-simulation}

\vspace{-2mm}
\mysubsection{Comparison to baseline.}
Our first set of experiments is designed to evaluate how spatial action maps compare to more traditional action representations.
To investigate this question, we ran experiments with an 18-dimensional steering commands action space, with actions representing taking a 25\,cm step forward in one of 18 possible turn directions.
We created a modified (baseline) version of our system by replacing the last layer of our network with a fully connected layer that outputs the predicted Q-value for each of the 18 actions. We also added two additional channels to the state representation to encode the relative position of every pixel location. This modified network mimics the DQN architectures and actions typical of other navigation algorithms \cite{kolve2017ai2,lowe2017multi,savva2017minos,anderson2018vision,wu2018building,xia2018gibson,yan2018chalet,savva2019habitat}, and yet is the same as ours in all other aspects.

Results are shown in Tab.~\ref{tab:action-representations}. We find that the steering commands baseline (right) is unable to learn effectively in any of the four environments. We conjecture the reasons are two-fold. First, the baseline network must learn a mapping from observations to discrete actions, which may be harder than the dense prediction enabled by spatial action maps. Second, the baseline agent can only reap rewards by executing a long sequence of short steps, and so it is difficult for the algorithm to achieve any reward at all in the early phases of training (Fig.~\ref{fig:curves-train}). In contrast, our method uses higher-level actions (that go directly to the selected location), and thus can discover rewards with fewer actions. This results in a policy that performs the task more completely and more efficiently (Fig.~\ref{fig:curves-eval}).

\mysubsection{Effect of shortest path movement primitive.}
To test the hypothesis that using a shortest path movement primitive helps our algorithm learn more efficiently, we ran experiments with a small modification to our system: the robot moves in a straight line to the selected target location (rather than along the shortest path). The results of this experiment (``No shortest path movement'' in Tab.~\ref{tab:action-representations}, and orange curve in Fig.~\ref{fig:curves-eval}) show a degradation in performance, particularly when there are more obstacles. Even though the shortest paths are computed with potentially inaccurate occupancy maps, it still seems to be advantageous to use them when navigating around obstacles.

\mysubsection{Effect of fixed step size.}
To test the hypothesis that movement primitives with longer trajectories help our algorithm learn more efficiently, we ran experiments with a different modification to our system: the length of any trajectory is fixed at 25\,cm (the same step size as the steering commands baseline).
The Q-network makes dense predictions as usual, but at each iteration, the agent steps a fixed distance in the direction of the position with the highest Q-value. The result of this variant (``Fixed step size'' in Tab.~\ref{tab:action-representations} and green curve in Fig.~\ref{fig:curves-eval}) shows that taking shorter steps indeed degrades performance. Even though it would be possible for the agent to take many small steps to achieve the same trajectory as a single long one, we find that it learns more quickly with longer trajectories. We conjecture that this could be due a less direct mapping from visual features to actions, as well as inconsistencies in Q-values predicted from different perspectives, causing the agent to waver between different endpoint targets as it takes many small steps.

\mysubsection{Effect of shortest path input channels.}
In addition to the overhead image, our system gives the agent three additional input image channels: (1) an image with a circle indicating the robot position, (2) an image with shortest path distances from the agent's position, and (3) an image with shortest path distances to the receptacle. To test whether the latter two of these channels are useful, we ran ablation studies for each of them.   The results (Tab.~\ref{tab:input-channels}) show that the channels provide little benefit in the environments with smaller obstacles, but help the system train more effectively in the most challenging environment (LargeDivider). We conjecture that providing the shortest path distances can help the agent prioritize target locations with objects that can be reached more easily (without having to go around large obstacles).

\mysubsection{Effect of shortest path partial rewards.} We train our agents with signed partial rewards given for pushing objects closer to or further away from the receptacle. We hypothesize that in environments with obstacles, it is important to give partial rewards based on changes in shortest path distance rather than Euclidean distance from the receptacle. We verify by running ablations that use Euclidean distance rather than shortest path distance, as shown in Tab.~\ref{tab:partial-rewards}. We observe that agents trained with Euclidean distance partial rewards indeed perform worse, particularly for the LargeDivider environment, where true shortest path distances to the receptacle can be much larger than Euclidean distances. We believe that giving partial rewards based on shortest path distances provides a better training signal to the agent. While Euclidean distances indicate whether an object is close to the receptacle, shortest path distances additionally factor in the need to push the object around the obstacles to get to the target receptacle. For completeness, we also show the performance when no partial rewards are given.

\begin{figure*}
\begin{center}
\begin{tabular}{@{}ccc}
\includegraphics[width=0.25\textwidth]{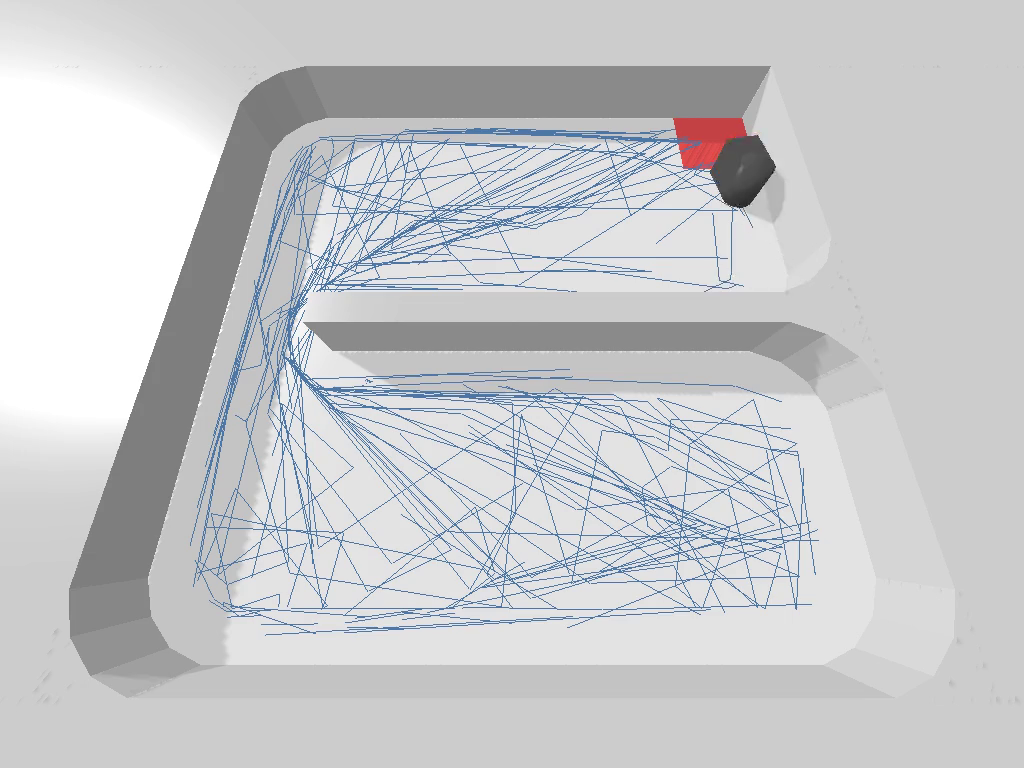} &
\includegraphics[width=0.25\textwidth]{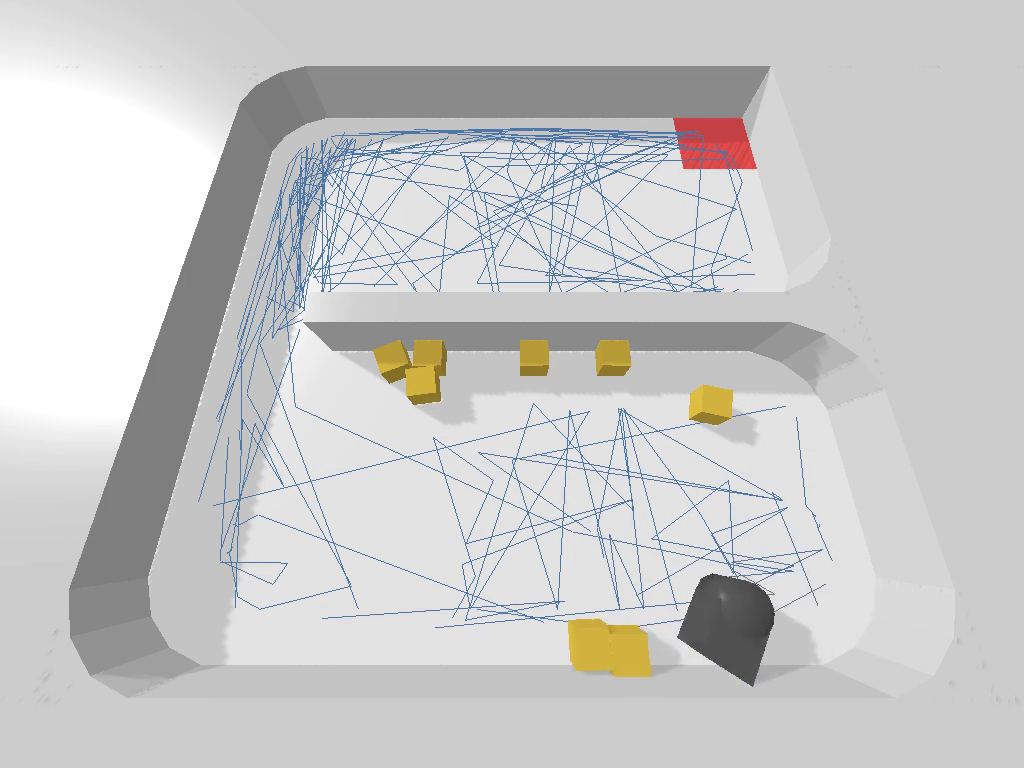} &
\includegraphics[width=0.25\textwidth]{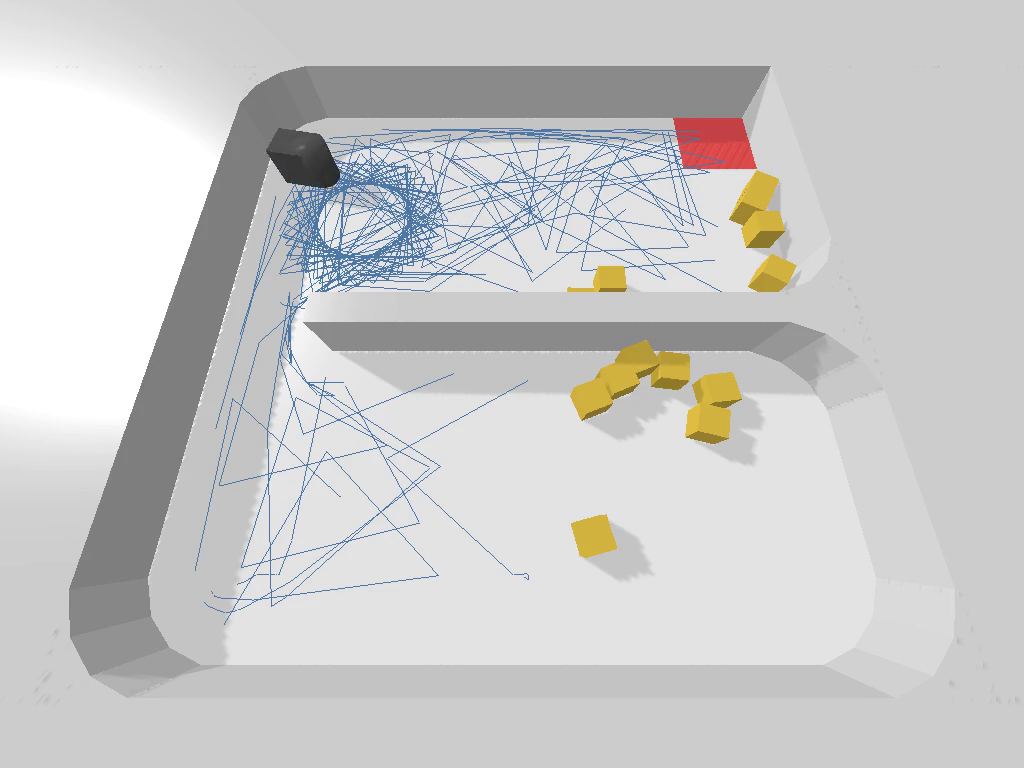} \\
\small{(a) Ours} & \small{(b) Ours w/o shortest path components} & \small{(c) Steering commands}
\end{tabular}
\end{center}
\vspace{-3mm}
\caption{\textbf{Episode trajectories.} In these runs on the LargeDivider environment, our method (a) follows a trajectory (blue line) that clears all objects efficiently. If we disable all shortest path components (b), it pushes objects straight towards the receptacle, and the ones in the bottom half get piled up against the divider. If we use the baseline action representation based on steering commands (c), the agent never learns to navigate effectively in this difficult environment.} 
\label{fig:episode-trajectories}
\vspace{-5mm}
\end{figure*}

\begin{table}
\caption{Effect of Shortest Path Input Channels}
\label{tab:input-channels}
\vspace{-3mm}
\begin{center}
\def\x{\hphantom{1}}
\begin{tabular}{l|ccc}
\toprule
\multirow{2}{*}{Environment} & \multirow{2}{*}{Ours} & No shortest path & No shortest path \\
                             &                       & from agent       & to receptacle \\
\midrule
SmallEmpty   & \x \bf{9.91} $\pm$ 0.11 & \x 9.82 $\pm$ 0.21 & n/a \\
SmallColumns & \x 9.18 $\pm$ 0.14 & \x 9.18 $\pm$ 0.32 & \x \bf{9.20} $\pm$ 0.22 \\
LargeColumns &   18.29 $\pm$ 0.45 &   18.40 $\pm$ 0.88 &   \bf{18.88} $\pm$ 0.49 \\
LargeDivider &   \bf{18.23} $\pm$ 0.92 &   16.87 $\pm$ 1.97 &   16.71 $\pm$ 1.49 \\
\bottomrule
\end{tabular}
\end{center}

\caption{Effect of Shortest Path Partial Rewards}
\label{tab:partial-rewards}
\vspace{-3mm}
\begin{center}
\def\x{\hphantom{1}}
\begin{tabular}{l|ccc}
\toprule
\multirow{2}{*}{Environment} & \multirow{2}{*}{Ours} & No shortest path & No partial \\
                             &                       & in partial rewards & rewards \\
\midrule
SmallEmpty   & \x \bf{9.91} $\pm$ 0.11 & n/a & \x 8.87 $\pm$ 1.61 \\
SmallColumns & \x \bf{9.18} $\pm$ 0.14 & \x 9.13 $\pm$ 0.28 & \x 9.02 $\pm$ 0.66 \\
LargeColumns &   \bf{18.29} $\pm$ 0.45 &   17.89 $\pm$ 0.97 & 13.95 $\pm$ 3.19 \\
LargeDivider &   \bf{18.23} $\pm$ 0.92 &   16.87 $\pm$ 1.97 & 12.02 $\pm$ 1.08 \\
\bottomrule
\end{tabular}
\end{center}

\caption{Effect of Removing All Shortest Path Components}
\label{tab:nosp}
\vspace{-3mm}
\setlength\tabcolsep{0.25em}
\begin{center}
\def\x{\hphantom{1}}
\begin{tabular}{l|cc|cc}
\toprule
\multirow{2}{*}{Environment}  &  \multirow{2}{*}{Ours}  &  Ours   & \multirow{2}{*}{Steering}  & Steering   \\
             &        & no shortest path &     & no shortest path\\
\midrule
SmEmpty   & \x \bf{9.91} $\pm$ 0.11 & \x 9.82 $\pm$ 0.10 & 1.38 $\pm$ 0.20 & 1.23 $\pm$ 0.69 \\
SmColumns & \x \bf{9.18} $\pm$ 0.14 & \x 8.05 $\pm$ 0.29 & 0.82 $\pm$ 0.33 & 0.72 $\pm$ 0.44 \\
LgColumns &   \bf{18.29} $\pm$ 0.45 &   15.63 $\pm$ 1.17 & 1.20 $\pm$ 0.64 & 0.33 $\pm$ 0.20 \\
LgDivider &   \bf{18.23} $\pm$ 0.92 &   10.06 $\pm$ 0.89 & 4.14 $\pm$ 2.21 & 0.26 $\pm$ 0.12 \\
\bottomrule
\end{tabular}
\end{center}
\vspace{-7mm}
\end{table} 

\mysubsection{Effect of removing all shortest path components.}  Here we remove all shortest path components of our system, specifically (1) shortest path movement primitive, (2) shortest path channels, and (3) shortest path partial rewards. We replace them with their straight-line variants: (1) straight-line movement, (2) channel containing Euclidean distance to receptacle, and (3) Euclidean distance partial rewards. The results are shown in Tab.~\ref{tab:nosp}. Indeed we see that for the SmallEmpty environment, there is no significant difference because there are no obstacles present. However, in more difficult environments (LargeDivider), we see that our method is much better at handling the obstacles. The difference can be seen clearly in the example trajectories visualized in Fig.~\ref{fig:episode-trajectories}. Our method pushes objects efficiently along shortest paths trajectories through free space (left). In contrast, the ablative version without shortest path components (middle) is less adept at navigating around obstacles and continually pushes objects up against the divider.

We similarly run the same ablations for the steering commands baseline. Specifically, we remove the shortest path channels and shortest path partial rewards, and replace them with straight-line variants. We find that while the baseline has some ability to handle obstacles when given shortest path channels and shortest path partial rewards, the performance on environments with obstacles can be dramatically worse without these shortest path components (Tab.~\ref{tab:nosp}).

\subsection{Real-World Experiments}

\begin{figure}
\vspace{2mm}
\includegraphics[width=\columnwidth]{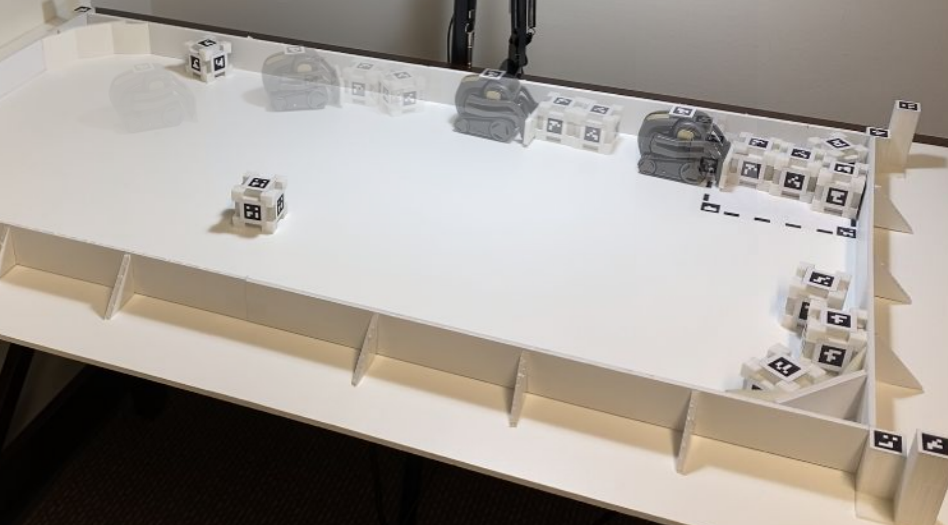} 
\vspace{-4mm}
\caption{\textbf{Emergent behaviors.} An Anki Vector robot executes a policy learned in simulation. It first pushes objects towards the wall, then uses the wall as a guide to simultaneously push several objects until they are all in the receptacle.}
\label{fig:emergent-behaviors}
\vspace{-5mm}
\end{figure}

We conduct experiments on the physical Anki Vector robots by replicating the SmallEmpty simulation environment on a tabletop. Our setup can be seen in Fig.~\ref{fig:emergent-behaviors}. We mount a camera over the tabletop, and affix fiducial markers to the robot and the objects, as well as the corners of the room. Using the overhead camera, we obtain real-time millimeter-level pose estimates of the robots and objects, which we then map into our simulation environment.

In this way, we enable our simulation environment to mirror the real-world environment. This means we can test our policies, which were trained in simulation, directly in the real-world setup. Given a state representation rendered by the simulation, our trained policy outputs a high-level action, which is executed on the physical robot by a low-level controller.
Overall, we find that trained agent behavior in the real-world environment is qualitatively similar to the simulation, but not quite as efficient due to differences in physical dynamics. We tested our best model on the SmallEmpty environment, and averaged across 5 test episodes, the real robot is able to push 8.4 out of 10 objects into the receptacle within 15 minutes, and all 10 within 30 minutes.
We show videos of our robot executing policies (that were learned in simulation) in the real-world environment at \url{https://spatial-action-maps.cs.princeton.edu}.

Looking at the videos, it is interesting to observe the emergent behaviors learned by the robot. Perhaps the most common is to first push objects up against the walls, and then later ``sweep'' multiple objects together along a wall with a single long trajectory ending at the receptacle (note the long paths along the walls in Fig.~\ref{fig:episode-trajectories}). This behavior is depicted in Fig.~\ref{fig:emergent-behaviors}, where a sequence of actions from one episode is shown in time-lapse as the robot pushes two objects at once (four near the end). Other emergent behaviors include retrying, where the the robot makes multiple attempts to nudge objects towards the receptacle after initial failure.

\subsection{Limitations}
\vspace{-2mm}
\mysubsection{Limitations of our setting and experiments.}
Because of our sim-to-real setup, our results are limited by
the accuracy of our simulations (including the measurement accuracy of physical properties in our setup), the quality and robustness of our motion primitive implementations, and our assumptions of perfect localization and mapping.
The first two of these are inherent to any system that is trained in simulation, and could be addressed by fine-tuning the learned model in the physical setup.
How uncertainty in state and observations should be incorporated into planning is an active research topic, particularly for deep reinforcement learning, and is orthogonal to our investigation into action space representations.

\mysubsection{Limitations of spatial action maps.}
One inherent limitation of spatial action maps is the use of high-level motion primitives, which assume that lower-level control can be handled separately.
Concretely, we output a destination that the agent should attempt to reach, and possibly what it should attempt to do once it gets there, and rely on a motion primitive to make that happen.
Other action representations, in contrast, may integrate learning of long-range planning and low-level control end-to-end.

Another potential limitation of spatial action maps is that they are only able to represent actions composed of navigating to a spatial location and performing a discrete action there. While this can be extended to higher dimensional settings (e.g., navigation in 2.5D multi-floor buildings) by using a set of 2D images or 3D grids, there may be tasks which are not compatible with this action representation.

\section{Conclusion}
In this work, we propose ``spatial action maps'' as a new action representation for mobile manipulation.
We study how best to use them in our experimental setting, and show that policies utilizing spatial action maps can be trained much more efficiently than with traditional action representations.
We also demonstrate that our agents trained in simulation can transfer directly to a real-world environment without further fine-tuning.
This work represents one step in a broader investigation of possible action representations, in one application domain.
In future work, it would be interesting to study other related action spaces (e.g., ones that mix navigation and manipulation) and other application domains (e.g., autonomous driving).

\section*{Acknowledgments}

The authors would like to thank Naveen Verma, Naomi Leonard, Anirudha Majumdar, Stefan Welker, and Yen-Chen Lin for fruitful technical discussions, as well as Julian Salazar for hardware support. This work was supported in part by the Princeton School of Engineering, as well as the National Science Foundation under IIS-1617236, IIS-1815070, and DGE-1656466.

\newpage
\bibliographystyle{plainnat}
\bibliography{references,navigation}

\end{document}